\documentclass[letterpaper]{article} 
\usepackage{aaai25}  
\usepackage{times}  
\usepackage{helvet}  
\usepackage{courier}  
\usepackage[hyphens]{url}  
\usepackage{graphicx} 
\usepackage{natbib}  
\usepackage{caption} 
\usepackage{algorithm}
\usepackage{algorithmic}
\usepackage{multirow}
\usepackage{amsmath}
\usepackage{amssymb}
\usepackage{amsthm}
\usepackage[table]{xcolor}
\usepackage{xcolor}
\usepackage{tcolorbox}
\usepackage{colortbl}
\usepackage{booktabs}
\usepackage{amsfonts}
\newtheorem{theorem}{Theorem}
\newtheorem{lemma}[theorem]{Lemma}
\newtheorem*{remark}{Remark}
\usepackage{array}
\usepackage[bookmarks=false]{hyperref}
\hypersetup{
    colorlinks=true,
    urlcolor=blue,
    citecolor=cyan
}
\usepackage{newfloat}
\usepackage{listings}
\DeclareCaptionStyle{ruled}{labelfont=normalfont,labelsep=colon,strut=off} 
\floatstyle{ruled}
\newfloat{listing}{tb}{lst}{}
\floatname{listing}{Listing}
\title{Fast and Slow Gradient Approximation for Binary Neural Network Optimization}
    
\author {
    Xinquan Chen\textsuperscript{\rm 1},
    Junqi Gao\textsuperscript{\rm 1,2},
    Biqing Qi\textsuperscript{\rm 2}\thanks{Corresponding author},
    Dong Li\textsuperscript{\rm 1,2},
    Yiang Luo\textsuperscript{\rm 1},
    Fangyuan Li\textsuperscript{\rm 1},
    Pengfei Li\textsuperscript{\rm 1}
}
\affiliations {
    \textsuperscript{\rm 1}Harbin Institute of Technology, Harbin, P.R.China,\\
    \textsuperscript{\rm 2}Shanghai Artificial Intelligence Laboratory, Shanghai, P.R.China\\
    \{xinquanchen0117,gjunqi97,qibiqing7\}@gmail.com,\\
    \{arvinlee826,normanluo668,jacklee19900212,lipengfei0208\}@gmail.com
}

\begin{document}

\maketitle

\begin{abstract}
Binary Neural Networks (BNNs) have garnered significant attention due to their immense potential for deployment on edge devices. However, the non-differentiability of the quantization function poses a challenge for the optimization of BNNs, as its derivative cannot be backpropagated. To address this issue, hypernetwork based methods, which utilize neural networks to learn the gradients of non-differentiable quantization functions, have emerged as a promising approach due to their adaptive learning capabilities to reduce estimation errors.
However, existing hypernetwork based methods typically rely solely on current gradient information, neglecting the influence of historical gradients. This oversight can lead to accumulated gradient errors when calculating gradient momentum during optimization. To incorporate historical gradient information, we design a Historical Gradient Storage (HGS) module, which models the historical gradient sequence to generate the first-order momentum required for optimization. To further enhance gradient generation in hypernetworks, we propose a Fast and Slow Gradient Generation (FSG) method. Additionally, to produce more precise gradients, we introduce Layer Recognition Embeddings (LRE) into the hypernetwork, facilitating the generation of layer-specific fine gradients.
Extensive comparative experiments on the CIFAR-10 and CIFAR-100 datasets demonstrate that our method achieves faster convergence and lower loss values, outperforming existing baselines.Code is available at \href{https://github.com/two-tiger/FSG}{github.com/FSG}.
\end{abstract}

%

\section{Introduction}
In recent years, Deep Neural Networks (DNNs) have demonstrated remarkable achievements across various computer vision tasks \cite{girshick_rich_2014, he_deep_2016, lecun_deep_2015, everingham_pascal_2015}. However, the increasing number of layers and parameters in deep learning models poses significant challenges. These challenges often result in substantial model sizes and elevated computational demands, hindering deployment on resource-constrained devices such as smartphones, cameras, and drones. 
To address these limitations, a plethora of compression techniques have emerged, including network pruning \cite{ding_resrep_2021, luo_thinet_2017}, low-rank approximation \cite{denton_exploiting_2014, hayashi_exploring_2019}, architectural redesign \cite{chen_addernet_2020, iandola_squeezenet_2016}, and network quantization \cite{banner_scalable_2018, ajanthan_mirror_2021}. Among these, network quantization—which employs reduced-precision data types for weight storage and computations—holds significant promise. For instance, Binary Neural Networks (BNNs), which restrict weights to binary values ($-1$ and $+1$) \cite{courbariaux2015binaryconnect}, can achieve remarkable efficiency by saving approximately 32 times in memory and boosting inference speed by a factor of 58 compared to full-precision networks \cite{rastegari2016xnor}.

Due to the non-differentiable nature of the sign function, conventional backpropagation is hindered in updating gradients for BNNs. BinaryConnect \cite{courbariaux2015binaryconnect} addresses this challenge by introducing the Straight-Through Estimator (STE), a technique used to approximate the gradient of the sign function during backpropagation. This method has become the fundamental paradigm for BNN optimization due to its effectiveness.
However, a significant issue remains: the discrepancy between the binary quantization applied during forward propagation and the gradient estimation during backpropagation leads to a gradient mismatch. This mismatch introduces inherent estimation errors, which impede the convergence and optimization of BNNs. As gradient propagation continues, these errors accumulate, further limiting the effectiveness of BNNs optimization.

To address the aforementioned challenges, a variety of methods have been developed. These include redesigned straight-through estimators \cite{qin2020forward, xu2021learning, wu2023estimator, lin2020rotated} and hypernetwork based methods, which utilize differentiable neural networks \cite{chen2019metaquant, liu2020quantnet} to directly learn the derivatives of non-differentiable functions, converting non-differentiable parts into differentiable operations for optimization.
However, the addition of neural networks introduces more complex operations and higher computational costs, which has led to the potential of hypernetwork based methods being overlooked. In fact, the ultimate goal of BNNs is to deploy on resource limited devices, and the training process for obtaining a quantized network version is not constrained by the computational limitations of the target device. Moreover, despite careful design, straight-through estimators cannot perfectly approximate the derivative of the binary thresholding function, resulting in inherent estimation errors and gradient mismatch.
Hypernetwork based methods offer a solution by dynamically adjusting the gradient approximation during training, thereby continually reducing estimation errors. This capability allows BNNs to optimize more flexibly and achieve higher performance.
Nonetheless, the current hypernetwork based methods \cite{chen2019metaquant} use the current weight gradient and the full precision weight to generate the final gradient. This method of generating gradients uses limited information and ignores the guiding role of historical gradients. Yet the optimization objective of the hypernetwork aligns with that of the final task, rather than the objective of generating a more suitable weight gradient, and does not match the optimization goal of generating a more appropriate weight gradient.

Motivated by the concept of momentum in Stochastic Gradient Descent with Momentum (SGD-M), which uses historical gradients to inform the current gradient, we have designed a Historical Gradient Storage (HGS) module. Momentum stabilizes the direction of gradient descent and accelerates optimization. Our HGS module stores a portion of historical gradients, treating them as time-series data that encapsulate gradient change information. These historical gradient sequences are then input into the hypernetwork to generate weight gradients.
Compared to using only the current gradient for learning, incorporating historical gradients provides additional context, enabling the network to generate gradients that more appropriate for BNN optimization. This approach enhances the optimization process for BNNs, facilitating more effective and efficient training.

To generate gradients more precisely, we have designed the Fast and Slow gradient Generation (FSG) method, inspired by the optimization method of SGD-M. This method employs two hypernetworks, referred to as fast-net and slow-net.
Slow-net utilizes models with strong long-sequence modeling capabilities, such as Mamba \cite{mamba} and LSTM \cite{6795963}, to capture the essence of historical gradient sequences in order to generate gradients consistent with the concept of momentum. This process requires summarizing accumulated historical data, similar to slow evolution. While fast-net utilizes a Multi-Layer Perceptron (MLP) to extract high-dimensional features from the current gradient, which can quickly generate the current weight gradient. The final gradient is obtained by combining these two components. Gradient generation is layer-specific, so in order to further guide slow-net to generate appropriate momentum, we add Layer Recognition Embeddings (LRE). Specifically, we add a learnable layer embedding vector as the recognition of layer information.
Compared with previous methods using a single hypernetwork, FSG enriches the process of gradient generation, allowing historical gradients to participate in gradient generation and providing richer information for gradient generation.

We conduct comprehensive experiments on the CIFAR-10 and CIFAR-100 datasets. The results demonstrate that our method outperforms various competitive benchmarks.
In summary, our contributions are as follows:

\begin{itemize}
    \item 
    Inspired by the idea of momentum, we examine the impact of historical gradients on gradient generation. To this end, we propose a HGS module for storing historical gradient sequences. By modeling these sequences, we generate gradient momentum, which guides the network in producing more suitable gradients.
   
    \item 
    We design a FSG method to guide the gradient generation process, fully leveraging the information from both historical and current gradients to produce more refined gradients. Additionally, we propose LRE to assist the hypernetwork in learning layer-specific gradients.
    \item 
    Extensive experiments on the CIFAR-10 and CIFAR-100 datasets demonstrate that our proposed FSG method achieves superior performance and faster convergence.
\end{itemize}

\section{Related Work}
The origins of BNNs can be traced back to Courbariaux's groundbreaking work \cite{courbariaux2016binarized}, where they employ a sign function to binarize network weights. To address the challenge of the sign function's non-differentiability during backpropagation, BinaryConnect \cite{courbariaux2015binaryconnect} introduces the STE as an effective approximation technique. The outstanding performance of STE has since become the cornerstone of optimization strategies in this field.

Later work primarily focuses on constraining the key aspect of BNN optimization: the derivative of the sign function. 
For instance, IR-Net \cite{qin2020forward} employs an Error Decay Estimator (EDE) in a two-stage process to gradually approximate the sign function and reduce gradient error. RBNN \cite{lin2020rotated} introduces a training-aware approximation of the sign function for gradient computation during backpropagation. FDA \cite{xu2021learning} utilizes combinations of sine functions in the Fourier frequency domain to estimate the gradient of the sign function during BNN training and employe neural networks to learn the approximation error. ReSTE \cite{wu2023estimator} leverages power functions to design a correction estimator that balances estimation error and gradient stability.

Despite efforts to design approximation functions, inherent estimation errors persist between these approximations and the sign function's derivatives, which are fundamentally uneliminable. Consequently, some recent approaches employ hypernetworks to transform the non-differentiable components of binary neural networks into differentiable operations.
For instance, MetaQuant \cite{chen2019metaquant} proposes incorporating a neural network into the backpropagation process to learn the non-differentiable gradients. Similarly, QuantNet \cite{liu2020quantnet} utilizes a differentiable subnetwork to directly binarize full-precision weights, avoiding reliance on the STE or any learnable gradient estimator.
Hypernetwork based methods \cite{andrychowicz2016learning} offer the advantage of adaptively adjusting gradient approximations during training, thereby reducing approximation errors. Unlike manually designed STE, these methods can dynamically refine the gradient approximations. However, they are limited in their ability to generate precise gradients. While the network learns based on the final loss, the optimization objective of hypernetworks is to generate better gradients for the sign function, leading to differing optimization goals.  

\section{Preliminaries}
\subsection{Definition of BNNs}
BNNs are often derived from fully accurate networks \cite{rastegari2016xnor}. For a neural network model \( f \) with \( n \) layers, parameterized by \(\mathcal{W} = [W_1, W_2, \ldots, W_n]\), where \( W_i \) represents the weight of the \(i\)-th layer, we define a pre-processing function \(\mathcal{A}(\cdot)\) and a quantization function \( Q(\cdot) \) to binarize the weights. This article investigates the pre-processing and quantization methods of DoReFa \cite{zhou2016dorefa}. The pre-processing function \(\mathcal{A}(\cdot)\) normalizes the weights, while the quantization function \( Q(\cdot) \) converts each value in the weight matrix to \(\{-1, +1\}\).
We have,
\begin{align}
    \hat{\mathcal{W}}&=\mathcal{A}(\mathcal{W})=\frac{\tanh(\mathcal{W})}{2\text{max}(|\tanh(\mathcal{W})|)}+\frac12, \\
    \mathcal{W}_b&=Q(\hat{\mathcal{W}})=2\frac{\text{round}\left[(2^k-1)\hat{\mathcal{W}}\right]}{2^k-1}-1.
\end{align}
For a training set \(\{\mathbf{x}, \mathbf{y}\}\) with \(\mathbf{N}\) instances, we incorporate quantization in the forward process. Consequently, the loss function for training a binary neural network is given by \(\ell(f(Q(\mathcal{A}(\mathcal{W})); \mathbf{x}), \mathbf{y})\). During the backpropagation process, we compute the derivative of \(\ell\) with respect to \(\mathcal{W}\),
\begin{equation}
    \frac{\partial\ell}{\partial\mathcal{W}}=\frac{\partial l}{\partial Q(\mathcal{A}(\mathcal{W}))}\frac{\partial Q(\mathcal{A}(\mathcal{W}))}{\partial\mathcal{A}(\mathcal{W})}\frac{\partial \mathcal{A}(\mathcal{W})}{\partial\mathcal{W}},
\end{equation}
where \(\frac{\partial Q(\mathcal{A}(\mathcal{W}))}{\partial \mathcal{A}(\mathcal{W})}\) is the Dirac delta function. This introduces a discontinuity in the backpropagation process, causing it to be interrupted at this point.

\subsection{Hypernetwork Based Method}
Hypernetwork based methods \cite{chen2019metaquant} introduce a shared neural network $\mathcal{M}$, which is trained simultaneously with the binary neural network. During the backpropagation process, the hypernetwork $\mathcal{M}$ receives two inputs, $g_{\mathcal{W}}=\frac{\partial \ell}{\partial \mathcal{W}}$ and $\hat{\mathcal{W}}$, and generates the derivative of $\ell$ to the quantization function $Q(\mathcal{A}(\cdot))$, i.e.
\begin{equation}
    \frac{\partial \ell}{\partial \mathcal{A}(\cdot)}=\mathcal{M}_{\phi}(g_{\mathcal{W}_b},\hat{\mathcal{W}}),
\end{equation}
the pre-processing function is differentiable, so the final derivative is,
\begin{equation}
    \frac{\partial\ell}{\partial\mathcal{W}}=\mathcal{M}_{\phi}(g_{\mathcal{W}},\hat{\mathcal{W}})\frac{\partial \mathcal{A}(\mathcal{W})}{\partial\mathcal{W}},
\end{equation}
please note that in the first iteration, the hypernetwork is not working as the gradient has not yet been generated. At this point, the STE function will be temporarily used as a substitute.

\begin{figure*}[t]
    \centering
    \includegraphics[width=0.7\textwidth]{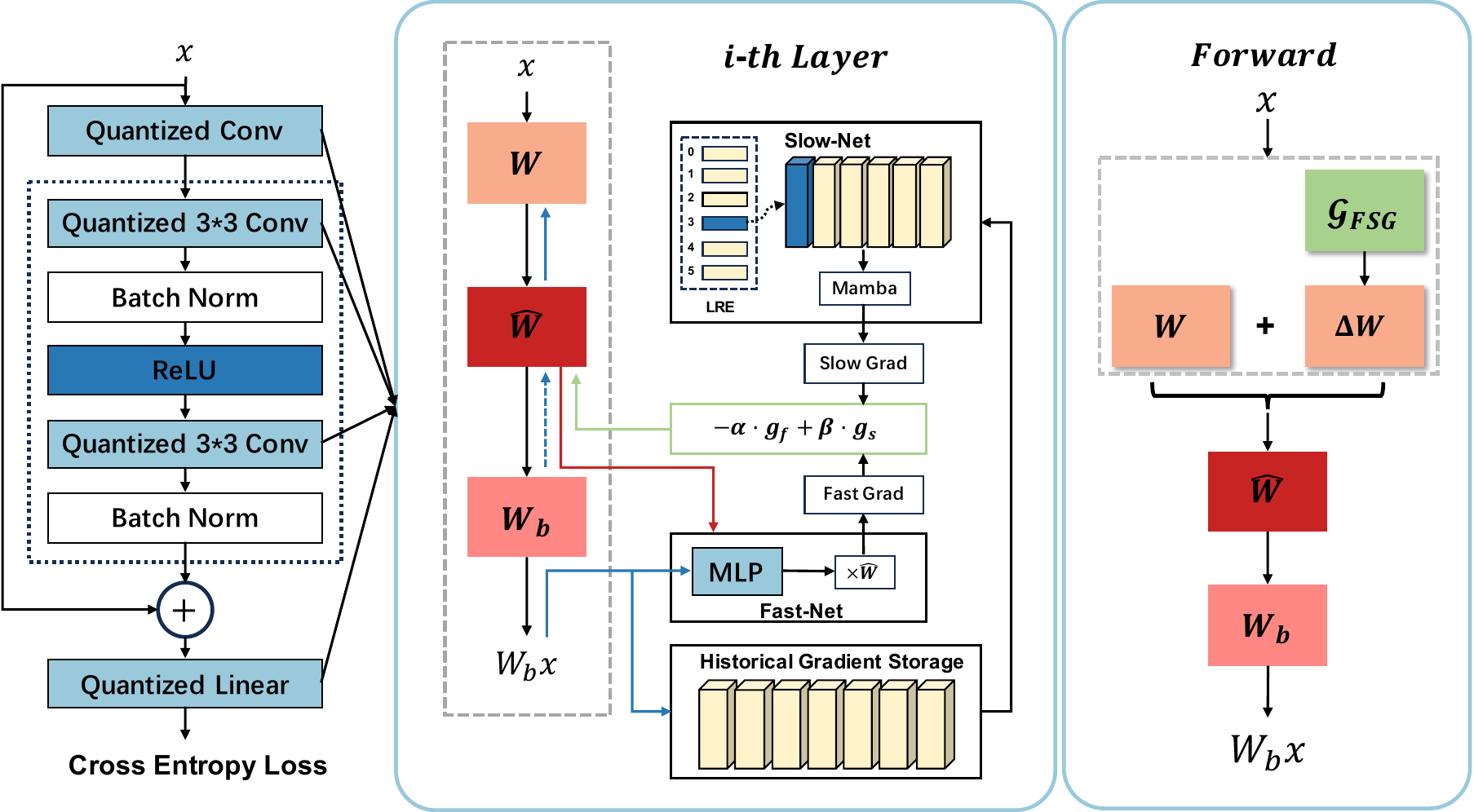}
    \caption{Fast and Slow Gradient Generation Illustration. Take ResNet as an example. During the backpropagation, the weight gradients from the previous iteration step are fed into the HGS and fast-net. Fast-net uses MLP to learn the scale of the weights, thereby obtaining the fast grad. The slow-net receives the historical gradient sequence from HGS and adds a LRE vector at the front of the sequence, then uses the mamba block to generate the slow grad. Ultimately, the slow grad and fast grad are combined through a weighted sum to generate the final gradients, replacing the non-differentiable parts (indicated by the blue dashed arrows). The forward process will be explained in Section Training of FSG.}
    \label{fig:1}
\end{figure*}

\section{Methodology}
\subsection{Motivation: Optimization Requires Historical Gradients}
Gradient descent, a cornerstone optimization algorithm as described in \cite{saad1998online}, relies on the gradient of the objective function with respect to the parameters. However, its reliance on instantaneous gradient can lead to premature convergence in local minima, hindering escape. To address this limitation, the momentum method incorporates exponential weighting from moving averages. By averaging past and current gradients with exponentially decaying weights, the method imparts inertia to the update direction. This inertia helps the algorithm to escape local minima more effectively, leading to improved convergence.
SGD-M has the following update formula,
\begin{equation}
    \begin{cases}
        \mathbf{v}_{t+1}=\beta\mathbf{v}_t-\alpha\nabla f(\mathbf{x}_t),\\
        \mathbf{x}_{t+1}=\mathbf{x}_t+\mathbf{v}_{t+1},
    \end{cases}
    \label{formual:m}
\end{equation}
where \(\mathbf{v}_{t+1}\) is the momentum, \(\beta\) is the momentum constant, \(\alpha\) is the step size, also known as the learning rate, \(\nabla f(\mathbf{x}_t)\) is the gradient of \(f\) at \(\mathbf{x}_t\), and the momentum \(\mathbf{v}_{t+1}\) can be expanded to:
\begin{equation}
    \begin{aligned}
        \mathbf{v}_{t+1}& =\beta\mathbf{v}_t-\alpha\nabla f(\mathbf{x}_t)  \\
        &=\beta^2\mathbf{v}_{t-1}-\beta\alpha\nabla f(\mathbf{x}_{t-1})-\alpha\nabla f(\mathbf{x}_t) \\
        &=\beta^3\mathbf{v}_{t-2}-\beta^2\alpha\nabla f(\mathbf{x}_{t-2})-\beta\alpha\nabla f(\mathbf{x}_{t-1}) \\
        & \quad -\alpha\nabla f(\mathbf{x}_t) \dots \\
    \end{aligned}
\end{equation}
From the above expansion of momentum, we can see that momentum is a linear combination of historical gradients, which means that the direction and magnitude of historical gradients can guide the direction of the current update and the step size in optimization. Therefore, it can be considered that when optimizing non-differentiable functions, it is also possible to benefit from historical gradients. We focuses on the role of historical gradients and introduces a module for storing historical gradients, called the Historical Gradient Storage (HGS) module, which saves the most recent \(l\) step gradients.

We treat the saved historical gradient sequence as a time-series data containing gradient change information. Mamba is a state-space model with a selection mechanism that can effectively filter out noise in historical gradient sequences, efficiently integrate historical gradients and provide more accurate momentum predictions. Therefore, we use a shared Mamba block \cite{mamba} to complete the modeling of the gradient sequence of each layer. We explain in detail the reasons for choosing Mamba as the momentum generator in the Appendix based on the principles of Mamba, and compare the performance of different sequence modeling models as momentum generators in Ablation Experiment.

\subsection{Fast and Slow Gradient Generation Method}
\textbf{Historical Gradient Storage Module.} In order to save the historical gradient, we introduce a module for storing historical gradients, termed the HGS module, and denote as $\mathcal{H}=\{h_1, h_2,..., h_n \}$, where $h_i$ represents the historical gradient snapshot of the $i$-th layer. For simplicity, we assume that $n$ layers of convolutional layers need to be generated, and the weights of the neural network are represented as $\mathcal{W}=[W_1, W_2,..., W_n]$, where $W_i \in \mathbb{R}^{C_{out}^i \times C_{in}^i \times K^i \times K^i}$ is the weight of the $i$-th convolutional layer has a corresponding gradient $g_{W_i} \in \mathbb{R}^{C_{out}^i \times C_{in}^i \times K^i \times K^i}$. We flatten the gradient to one dimension and save it as $g_{w_i} \to \overline{g}_{w_i} \in \mathbb{R}^{\xi^i \times 1}$, where $\xi^i=C_{out}^i \cdot C_{in}^i \cdot K^i \cdot K^i$, to control memory usage, we only retain the gradients from the most recent $l$ steps,
\begin{equation}
    h_i^t=\text{Concat}[\overline{g}_{w_i}^{t-l+1},...,\overline{g}_{w_i}^{t-1},\overline{g}_{w_i}^{t}] \in \mathbb{R}^{\xi^i \cdot l \times 1},
\end{equation}
where $\text{Concat}$ denotes the concatenation along the appropriate dimension to form a vector of length $\xi^i \cdot l$,  $\overline{g}_{W_i}^{t}$ is the flattened gradient of the weight $W_i$ at step $t$. 

\noindent \textbf{Fast and Slow Gradient Generation Mechanism.} As shown in the Fig. \ref{fig:1}, we introduce two hypernetworks shared by each layer $\mathcal{M}_f$ and $\mathcal{M}_s$, named fast-net and slow-net respectively. Fast-net receives the full-precision weights \(\hat{W}_i^{t}\) of the previous step and the gradients \(g_{W_i}^{t}\) of the previous step to generate the gradients \(g_{\mathcal{W}}^{t+1}\) for updating. Slow-net receives the historical gradient sequence \(h_{i}^{t}\) provided by the HGS module to generate the gradient momentum \(\mathbf{v}_{t+1}\) for updating. Therefore, there is the following update form, for the \(i\)-th layer,
\begin{equation}
    \mathbf{W}_i^{t+1}=\mathbf{W}_i^t-\alpha\mathcal{M}_f(g_{W_i}^{t}, \hat{W}_i^{t})+\beta\mathcal{M}_s(h_{i}^{t})
    \label{corporate}
\end{equation}
Slow-net is responsible for modeling historical gradient sequences to generate gradients, which is consistent with the gradient momentum composed of historical gradients. The generation of momentum requires the accumulation of historical information, similar to a slow evolution. Fast-net, on the other hand, learns the high-dimensional features of the current gradient and is a fast gradient generation method.

\noindent \textbf{Layer Recognition Embedding.} To enable the Mamba block to discern the layer-specific information and prevent the gradients from confusing each other during the training process, inspired by ViM \cite{vim} and BERT \cite{devlin2018bert}, we propose LRE, which initializing a learnable embedding vector for each layer of the model $\mathbb{E} \in \mathbb{R}^{n \times d}$, $d$ is the dimension of the embedding space. For a given layer index $i$, LRE will return the corresponding embedding vector, which we define as $\mathbf{t}_{i} \in \mathbb{R}^{d}$. Define $\hbar = [\hbar_1, \hbar_2,...,\hbar_n]$, where $\hbar_i$ is the input vector formed by concatenating the vector $\mathbf{t}_{i}$ with the projected historical gradient.
\begin{equation}
    \begin{aligned}
    \hbar_i^t&=[\mathbf{t}_{i};\overline{g}_{w_i}^{t-l}\mathbf{W}_a,...,\overline{g}_{w_i}^{t-1}\mathbf{W}_a,\overline{g}_{w_i}^{t}\mathbf{W}_a] \\
    &= [\mathbf{t}_{i}; h_i\mathbf{W}_a] \in \mathbb{R}^{(\xi^i \cdot l + 1) \times d}
    \end{aligned}
\end{equation}
where $\mathbf{W}_a \in \mathbb{R}^{1 \times d}$ is a projection matrix that maps each flattened gradient $\overline{g}_{w_i}^{t-k}$ onto a $d$ dimensional space. Next, we input $\hbar_i^t$ into the Mamba block, we slice the output vector, take the last $\xi^i$ sequences, and project them to a 1-dimensional vector,
\begin{equation}
    \overline{g}_Q = \text{Mamba}(\hbar_i^t)[:, -\xi^i:, :]\mathbf{W}_b  \in \mathbb{R}^{\xi^i \times 1},
\end{equation}
where $\mathbf{W}_b \in \mathbb{R}^{d \times 1}$ is a projection matrix. We reshape the output vector $\overline{g}_Q \to g_Q \in \mathbb{R}^{C_{out}^i \times C_{in}^i \times K^i \times K^i}$ and obtain the derivative of quantized function. Compared to using only the current gradient information to generate gradients, this approach uses historical gradients to guide the generation of the current gradient, fully utilizing the information. 

\subsection{Training of FSG}
\label{forward}
Similar to \cite{chen2019metaquant}, we will incorporate shared fast-net $\mathcal{M}_f$ and slow-net $\mathcal{M}_s$ into the forward process of the model, as shown in Fig. \ref{fig:1}. Eq. \ref{corporate} is used to merge the gradients generated by FSG and the weight $W_i$ of $i$-th layer. Then we have the forward process in the $i$-th convolutional layer,

\begin{equation}
    \begin{aligned}
        \hat{W}_i^{t+1}&=\mathcal{A}(W_i^{t+1}) \\
        &=\mathcal{A}\left[W_i^{t}-\alpha\mathcal{M}_f(\mathcal{G}_{W_i}^{t}, \hat{W}_i^{t})\frac{\partial \mathcal{A}(W_i^t)}{\partial W_i^t}+\beta\mathcal{M}_s(\hbar_{i}^{t})\right]
    \end{aligned}
\end{equation}
At the same time, we register the generated gradient \(\mathcal{G}_{FSG} = \beta\mathcal{M}_s(\hbar_{i}^{t})-\alpha\mathcal{M}_f(\mathcal{G}_{W_i}^{t}, \hat{W}_i^{t})\frac{\partial \mathcal{A}(W_i^t)}{\partial W_i^t}\) in the optimizer to replace $g_{\mathcal{W}}=\frac{\partial \ell}{\partial \mathcal{W}}$, and we calculate the cross entropy loss $\mathcal L$ based on the results generated by quantifying weights:
\begin{equation}
    \mathcal L=\ell\left(f(Q(\hat{W}_i^{t+1}); \mathbf{x} ),\mathbf{y}\right),
\end{equation}
therefore, $\mathcal{M}_f$ and $\mathcal{M}_s$ are associated with the final loss and can be used for backpropagation to update parameters. We assume that $\mathcal{M}_f$ and $\mathcal{M}_s$ are parametered by $\phi_f$ and $\phi_s$, respectively, during the backpropagation process,
\begin{equation}
    \frac{\partial\ell}{\partial\phi_{f}^{t+1}}=\frac{\partial\ell}{\partial\hat{\mathcal{W}}^{t+1}}\frac{\partial\hat{\mathcal{W}}^{t+1}}{\partial\phi_{f}^{t+1}}=-\alpha\cdot \mathcal{G}_{FSG}\frac{\partial \mathcal{M}_f(\mathcal{G}_{W_i}^{t}, \hat{W}_i^{t})}{\partial \phi_{f}^{t+1}}
    \label{gradient_f}
\end{equation}
\begin{equation}
    \frac{\partial\ell}{\partial\phi_{s}^{t+1}}=\frac{\partial\ell}{\partial\hat{\mathcal{W}}^{t+1}}\frac{\partial\hat{\mathcal{W}}^{t+1}}{\partial\phi_{s}^{t+1}}=\beta \cdot \mathcal{G}_{FSG}\frac{\partial \mathcal{W}^{t+1}}{\partial \mathcal{A}(\mathcal{W}^{t+1})}\frac{\partial \mathcal{M}_s(\hbar_t)}{\partial \phi_{s}^{t+1}}
    \label{gradient_s}
\end{equation}
After the hypernetwork updates, we use the registered gradient $\mathcal{G}_{FSG}$ to update all parameters with the optimizer.

\subsection{Convergence Analysis of FSG}
In this section, we analyze the convergence of the proposed fast and slow gradient generation mechanism under the general convex function condition. Randomly sample a dataset \(\mathcal{D}=\{(a_1,b_1),(a_2,b_2),\dots,(a_N,b_N)\}\). The task is to predict the label \(b\) given the input \(a\), i.e. determine an optimal function \(\phi\) that minimizes the expected risk \(\mathbb{E}[L(\phi(a),b)]\), where \(L(\cdot,\cdot)\) represents the loss function and the function \(\phi\) is the prediction function in a certain function space.

In practice, in order to reduce the range of the objective function, it is necessary to parameterize \(\phi(\cdot)\) to \(\phi(\cdot;x)\), where \(x\) is a parameter. Using empirical risk to approximate expected risk requires solving the following minimization problem:
\begin{equation}
    \min_x\quad\frac1N\sum_{i=1}^NL(\phi(a_i;x),b_i)=\mathbb{E}_{(a,b)\thicksim\hat{P}}[L(\phi(a;x),b)],
\end{equation}
define \(f_i(x)=L(\phi(a_i;x),b_i)\), then we only need to consider the following stochastic optimization problem:
\begin{equation}
    \min_{x\in\mathbb{R}^n}\quad f(x)\stackrel{\mathrm{def}}{=}\frac1N\sum_{i=1}^Nf_i(x)
    \label{formual:question}
\end{equation}

We give the following theorem:

\begin{theorem}
    (Convergence of FSG) Let formula \ref{corporate} run \(t\) iterations. By setting \(\alpha=\frac{C}{\sqrt{t+1}}\),
    \begin{equation}
        \begin{aligned}
            &\mathbb{E}[f(\widehat{\mathbf{x}}_t)-f(\mathbf{x}^*)] \\
            \leq &\frac\beta{(1-\beta)(t+1)}(f(\mathbf{x}_0)-f(\mathbf{x}^*)) +\frac{(1-\beta)\|\mathbf{x}_0-\mathbf{x}_*\|^2}{2C\Omega\kappa\sqrt{t+1}} \\
            & +\frac{C\Theta\rho(G^2+\delta^2)}{2\Omega\kappa(1-\beta)\sqrt{t+1}}\\
        \end{aligned}
    \end{equation}
    Where \(C, \Omega, \kappa,\Theta, \rho \) is a normal number, \(\mathbf{x}^*\) is the optimal solution and \(\widehat{\mathbf{x}}_t=\sum_{k=0}^t\mathbf{x}_k/(t+1)\).
\end{theorem}
\begin{remark}
    Theorem 1 describes the convergence of FSG, i.e. the expected error with the optimal solution decreases as the number of iterations increases for the iterative formula $t$ times, and the convergence order is $1/{\sqrt{t+1}}$. The convergence speed is slow at the beginning of the iteration. As the number of iterations increases, the convergence speed will increase. It can almost guarantee convergence.
\end{remark}
The proof is given in the Appendix.

\section{Experiment}
In this section, we follow the setting of \cite{chen2019metaquant} and explore whether FSG can effectively learn the derivatives of quantization functions on the CIFAR-10/100 dataset. Then we follow the setting of \cite{chen2019metaquant,qin2020forward} to compare it with several State-Of-The-Art (SOTA) methods.

\begin{table*}[ht]
\centering
\scriptsize
\renewcommand{\arraystretch}{0.62}
\caption{Results on CIFAR-10 dataset. Loss is the cross-entropy loss value. FP Acc is the result of the full precision version of the backbone on this dataset. The bold font is utilized to denote the optimal results, the underline is applied to indicate the suboptimal outcomes.}
\begin{tabular}{cccccccc}\toprule
\textbf{Backbone} & \textbf{Forward} & \textbf{Optimization} &  \textbf{Backward} & \textbf{Train Acc (\%)} & \textbf{Test Acc (\%)} & \textbf{Loss} & \textbf{FP Acc (\%)} \\ 
\midrule
\multirow{11}{*}{ResNet-20} & \multirow{11}{*}{DoReFa} & \multirow{5}{*}{SGD} & STE & 55.469 $\pm$ 3.342 & 46.81 $\pm$ 4.893 & 1.451 $\pm$ 0.395 & \multirow{11}{*}{91.70} \\
& & & FCGrad & \underline{92.084 $\pm$ 1.103} & \underline{88.54 $\pm$ 0.291} & \underline{0.220 $\pm$ 0.021} & \\
& & & LSTMFC & 92.070 $\pm$ 1.012 & 88.28 $\pm$ 0.810 & 0.228 $\pm$ 0.027 & \\
& & & \cellcolor{gray!20} \textbf{FSG(ours)} & \cellcolor{gray!20} \textbf{92.302} $\pm$ \textbf{1.247} & \cellcolor{gray!20} \textbf{88.67} $\pm$ \textbf{0.633} & \cellcolor{gray!20} \textbf{0.218} $\pm$ \textbf{0.012} & \\
\cmidrule{3-7}
& & \multirow{5}{*}{Adam} & STE & 93.138 $\pm$ 0.976 & 87.96 $\pm$ 0.173 & 0.151 $\pm$ 0.042 & \\
& & & FCGrad & 94.331 $\pm$ 0.986 & 89.96 $\pm$ 0.068 & 0.148 $\pm$ 0.027 & \\
& & & LSTMFC & \underline{94.768 $\pm$ 1.023} & \underline{89.98 $\pm$ 0.103} & \underline{0.148 $\pm$ 0.029} & \\
& & & \cellcolor{gray!20} \textbf{FSG(ours)} & \cellcolor{gray!20} \textbf{94.926} $\pm$ \textbf{0.747} & \cellcolor{gray!20} \textbf{91.00} $\pm$ \textbf{0.804} & \cellcolor{gray!20} \textbf{0.143} $\pm$ \textbf{0.039} & \\

\midrule
\multirow{11}{*}{ResNet-32} & \multirow{11}{*}{DoReFa} & \multirow{5}{*}{SGD} & STE & 55.469 $\pm$ 2.948 & 47.12 $\pm$ 4.683 & 1.443 $\pm$ 0.397 & \multirow{11}{*}{92.13} \\
& & & FCGrad & \underline{95.377 $\pm$ 1.588} & 89.93 $\pm$ 0.246 & 0.154 $\pm$ 0.024 & \\
& & & LSTMFC & 95.807 $\pm$ 1.643 & \underline{90.40 $\pm$ 0.149} & \underline{0.150 $\pm$ 0.021} & \\
& & & \cellcolor{gray!20} \textbf{FSG(ours)} & \cellcolor{gray!20} \textbf{96.898} $\pm$ \textbf{1.476} & \cellcolor{gray!20} \textbf{90.53} $\pm$ \textbf{0.192} & \cellcolor{gray!20} \textbf{0.145} $\pm$ \textbf{0.019} & \\
\cmidrule{3-7}
& & \multirow{5}{*}{Adam} & STE & \underline{97.056 $\pm$ 1.010} & 89.47 $\pm$ 0.097 & 0.124 $\pm$ 0.023 & \\
& & & FCGrad & 95.288 $\pm$ 1.648 & \underline{90.28 $\pm$ 0.068} & \underline{0.900 $\pm$ 0.018} & \\
& & & LSTMFC & 95.140 $\pm$ 1.779 & 90.15 $\pm$ 0.074 & 0.907 $\pm$ 0.023 & \\
& & & \cellcolor{gray!20} \textbf{FSG(ours)} & \cellcolor{gray!20} \textbf{97.632 $\pm$ 1.310} & \cellcolor{gray!20} \textbf{91.42} $\pm$ \textbf{0.140} & \cellcolor{gray!20} \textbf{0.083} $\pm$ \textbf{0.016} & \\

\midrule
\multirow{11}{*}{ResNet-44} & \multirow{11}{*}{DoReFa} & \multirow{5}{*}{SGD} & STE & 59.375 $\pm$ 3.322 & 49.73 $\pm$ 4.628 & 1.240 $\pm$ 0.081 & \multirow{11}{*}{93.55} \\
& & & FCGrad & 92.228 $\pm$ 2.029 & 89.93 $\pm$ 0.246 & 0.711 $\pm$ 0.024 & \\
& & & LSTMFC & \underline{94.548 $\pm$ 1.947} & \underline{90.40 $\pm$ 0.149} & \underline{0.723 $\pm$ 0.022} & \\
& & & \cellcolor{gray!20} \textbf{FSG(ours)} & \cellcolor{gray!20} \textbf{98.916} $\pm$ \textbf{1.398} & \cellcolor{gray!20} \textbf{91.91} $\pm$ \textbf{0.122} & \cellcolor{gray!20} \textbf{0.031} $\pm$ \textbf{0.015} & \\
\cmidrule{3-7}
& & \multirow{5}{*}{Adam} & STE & \underline{97.656 $\pm$ 0.996} & 90.21 $\pm$ 0.075 & 0.083 $\pm$ 0.012 & \\
& & & FCGrad & 90.584 $\pm$ 1.573 & \underline{90.98 $\pm$ 0.068} & \underline{0.083 $\pm$ 0.022} & \\
& & & LSTMFC & 90.147 $\pm$ 1.392 & 90.95 $\pm$ 0.074 & 0.082 $\pm$ 0.023 & \\
& & & \cellcolor{gray!20} \textbf{FSG(ours)} & \cellcolor{gray!20} \textbf{99.262} $\pm$ \textbf{1.322} & \cellcolor{gray!20} \textbf{92.78} $\pm$ \textbf{0.053} & \cellcolor{gray!20} \textbf{0.022} $\pm$ \textbf{0.011} & \\
\bottomrule
\end{tabular}
\label{cifar10-exp1}
\end{table*}

\begin{table*}[h]
\centering
\scriptsize
\renewcommand{\arraystretch}{0.62}
\caption{Results on CIFAR-100 dataset. Loss is the cross-entropy loss value. FP Acc is the result of the full precision version of the backbone on this dataset.The bold font is utilized to denote the optimal results, the underline is applied to indicate the suboptimal outcomes.}
\begin{tabular}{cccccccc}\toprule
\textbf{Backbone} & \textbf{Forward} & \textbf{Optimization} &  \textbf{Backward} & \textbf{Train Acc (\%)} & \textbf{Test Acc (\%)} & \textbf{Loss} & \textbf{FP Acc (\%)} \\ 
\midrule
\multirow{10}{*}{ResNet-56} & \multirow{10}{*}{DoReFa} & \multirow{5}{*}{SGD} & STE & 51.126 $\pm$ 7.374 & 42.27 $\pm$ 8.143 & 1.890 $\pm$ 0.178 & \multirow{11}{*}{71.22} \\
& & & FCGrad & 73.934 $\pm$ 1.639 & \underline{63.25 $\pm$ 0.935} & \underline{0.861 $\pm$ 0.012} & \\
& & & LSTMFC & \underline{73.934 $\pm$ 1.454} & 62.95 $\pm$ 2.183 & 0.901 $\pm$ 0.010 & \\
& & & \cellcolor{gray!20} \textbf{FSG(ours)} & \cellcolor{gray!20} \textbf{74.239} $\pm$ \textbf{1.691} & \cellcolor{gray!20} \textbf{64.02} $\pm$ \textbf{1.344} & \cellcolor{gray!20} \textbf{0.822} $\pm$ \textbf{0.012} & \\
\cmidrule{3-7}
& & \multirow{5}{*}{Adam} & STE & \underline{81.250 $\pm$ 0.642} & 64.78 $\pm$ 0.533 & 0.618 $\pm$ 0.032 & \\
& & & FCGrad & 79.944 $\pm$ 0.923 & \underline{66.56 $\pm$ 0.351} & \underline{0.684 $\pm$ 0.022} & \\
& & & LSTMFC & 79.336 $\pm$ 0.931 & 66.30 $\pm$ 0.793 & 0.686 $\pm$ 0.024 & \\
& & & \cellcolor{gray!20} \textbf{FSG(ours)} & \cellcolor{gray!20} \textbf{81.452} $\pm$ \textbf{0.873} & \cellcolor{gray!20} \textbf{69.48} $\pm$ \textbf{0.979} & \cellcolor{gray!20} \textbf{0.604} $\pm$ \textbf{0.019} & \\

\midrule
\multirow{10}{*}{ResNet-110} & \multirow{10}{*}{DoReFa} & \multirow{5}{*}{SGD} & STE & 53.159 $\pm$ 9.793 & 43.42 $\pm$ 8.902 & 1.783 $\pm$ 0.211 & \multirow{11}{*}{72.54} \\
& & & FCGrad & \underline{82.379 $\pm$ 1.553} & \underline{65.15 $\pm$ 2.490} & \underline{0.602 $\pm$ 0.023} & \\
& & & LSTMFC & 81.074 $\pm$ 1.457 & 64.75 $\pm$ 2.850 & 0.619 $\pm$ 0.027 & \\
& & & \cellcolor{gray!20} \textbf{FSG(ours)} & \cellcolor{gray!20} \textbf{84.733} $\pm$ \textbf{2.773} & \cellcolor{gray!20} \textbf{65.23} $\pm$ \textbf{0.822} & \cellcolor{gray!20} \textbf{0.601} $\pm$ \textbf{0.012} & \\
\cmidrule{3-7}
& & \multirow{5}{*}{Adam} & STE & 85.156 $\pm$ 1.344 & 66.84 $\pm$ 1.205 & 0.548 $\pm$ 0.016 & \\
& & & FCGrad & 85.299 $\pm$ 1.284 & \underline{68.74 $\pm$ 0.363} & \underline{0.483 $\pm$ 0.014} & \\
& & & LSTMFC & \underline{85.699 $\pm$ 1.273} & 67.14 $\pm$ 1.286 & 0.503 $\pm$ 0.013 & \\
& & & \cellcolor{gray!20} \textbf{FSG(ours)} & \cellcolor{gray!20} \textbf{85.738 $\pm$ 1.031} & \cellcolor{gray!20} \textbf{68.15} $\pm$ \textbf{0.973} & \cellcolor{gray!20} \textbf{0.462} $\pm$ \textbf{0.011} & \\
\bottomrule
\end{tabular}
\label{cifar100-exp1}
\end{table*}

\subsection{Experiment Setup}
In this work, we use CIFAR-10 and CIFAR-100 as benchmark datasets. CIFAR-10 \cite{krizhevsky2009learning} and CIFAR-100 \cite{krizhevsky2009learning} are commonly used datasets for image classification, consisting of 50000 training images and 10000 test images, with 10 and 100 categories respectively. Each image has a size of $32 \times 32$ and includes an RGB color channel. For CIFAR-10, we conduct experiments using ResNet-20/32/44 architecture, and for CIFAR-100, we conduct experiments using ResNet-56/110 architecture. For fair comparison, except for the first convolutional layer and the last fully connected layer, other layers are binarized in this experiment. See the Appendix for more experimental settings and details.

\subsection{Experimental Results and Analysis}
In this experiment, we aim to investigate whether the FSG can produce effective gradients. To this end, we compare FSG with the STE \cite{zhou2016dorefa}, FCGrad \cite{chen2019metaquant} and LSTMFC \cite{chen2019metaquant}. Our experiments are conducted on both SGD and Adam optimizers.

Tab. \ref{cifar10-exp1} shows the experimental results of STE, FCGrad, LSTMFC and FSG using different optimization methods on the CIFAR-10 dataset. The results show that the accuracy of FSG is significantly better than other types of hypernetworks, and it has a smaller loss, with ResNet-32 having the smallest difference of only 0.81\% compared to the results of full precision inference. 

Tab. \ref{cifar100-exp1} shows the experimental results of STE, FCGrad, LSTMFC, and FSG on the CIFAR-100 dataset using different optimization methods. Similar to CIFAR-10, FSG demonstrates superior performance over LSTMFC across a range of scenarios, achieving a higher accuracy rate of 68.04\% compared to LSTMFC's 66.30\%. This indicates that FSG's incorporation of historical gradient information significantly improves the optimization process by providing more accurate estimates of the gradients needed for learning quantization schemes.

\begin{figure*}[ht]
    \centering
    \includegraphics[width=\textwidth]{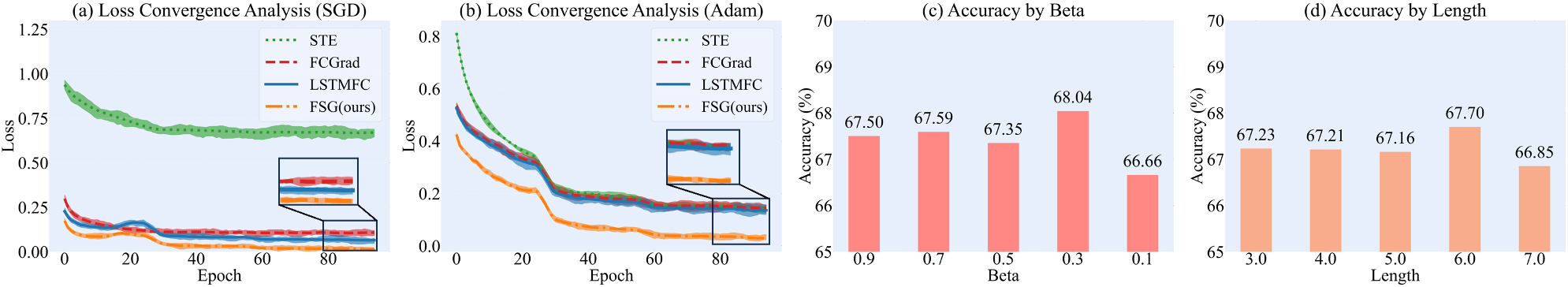}
    \caption{(a) Loss Curve of ResNet44 on CIFAR-10 Dataset with SGD Optimizer. (b) Loss Curve of ResNet44 on CIFAR-10 Dataset with Adam Optimizer. (c) Accuracy by Hyperparameter Beta. (d) Accuracy by Hyperparameter Length.}
    \label{fig:exp}
\end{figure*}

\subsection{Convergence Analysis}
In this comparative study, we aim to evaluate the convergence speed of different optimizers, namely STE, FCGrad, LSTMFC, and FSG, by analyzing their loss curves throughout the training process, as depicted in Fig \ref{fig:exp}. The loss curves are plotted to visualize the performance of each method during training, using the DoReFa quantization function and the ResNet44 network architecture on the CIFAR-10 dataset. The loss curves reveal that FSG exhibit superior convergence characteristics, with a notably faster descent rate and consistently lower loss values compared to FCGrad, LSTMFC and STE. Furthermore, compared to traditional SGD optimizer, the Adam optimizer exhibits a more stable and consistent optimization process, which reflects that our momentum generation gradient strategy is more suitable for the Adam optimizer.

\subsection{Ablation Experiment}
In this section, we investigate the effects of various sequence models as slow-nets on performance and examine the influence of the balancing parameter $\alpha$, which governs the blending of fast and slow gradients within the FSG method, alongside the impact of the hyperparameter $l$, which represents the length of the historical gradient memory fed into the slow network.

\textbf{Using different slow-net.} In this experiment, we use two sequence modeling models, including LSTM and Mamba, as slow-nets. For the CIFAR10 and CIFAR100 datasets, we employed ResNet-44 and ResNet-56 as the respective backbone architectures. As evident from Table \ref{table:slow}, utilizing Mamba as a momentum generator exhibited superior performance, indicating that Mamba possesses enhanced capabilities in modeling long sequences. Its distinctive selection mechanism was found to aid in the filtering of noise within the historical gradient sequence.

\begin{table}[h]
\centering
\scriptsize
\renewcommand{\arraystretch}{0.6}
\caption{Performance comparison of different sequence modeling models as slow-net.}
\begin{tabular}{ccccc}\toprule
\textbf{Backbone} & \textbf{Optimization} &  \textbf{Slow-net} & \textbf{Test Acc (\%)} & FP Acc (\%) \\ 
\midrule
\multirow{2}{*}{ResNet-44} & \multirow{2}{*}{CIFAR10} & LSTM & 92.07 $\pm$ 0.066 & \multirow{2}{*}{93.55} \\
\cmidrule{3-4}
& & Mamba & 92.63 $\pm$ 0.052 & \\
\midrule
\multirow{2}{*}{ResNet-56} & \multirow{2}{*}{CIFAR100} & LSTM & 67.60 $\pm$ 0.670 & \multirow{2}{*}{71.22} \\
\cmidrule{3-4}
& & Mamba & 68.04 $\pm$ 0.948 & \\
\bottomrule
\end{tabular}
\label{table:slow}
\end{table}

\textbf{Influence of $\beta$.} In this experiments, we explore the sensitivity of the FSG method to the combination parameter $\beta$, selecting values of 0.9, 0.7, 0.5, 0.3, and 0.1 for evaluation on the CIFAR-100 dataset. We utilize the ResNet56 architecture and the DoReFa quantization function for our analysis. From the result in the Fig. \ref{fig:exp}, it can be found that the highest accuracy is achieved when $\beta$ is set to 0.3. This indicates that the current gradient generated by fast-net dominates the update, which helps to respond quickly to changes in gradient and speed up convergence. At the same time, the gradient momentum generated by the historical gradient sequence can still reduce the gradient oscillation caused by noise to a certain extent.Based on these results, we recommend starting with a smaller value of $\beta$ (around 0.3).

\textbf{Influence of $l$.} For $l$, we select 3, 4, 5, 6, and 7 to conduct experiments on CIFAR-100, using ResNet56 as the network architecture and DoRefa as the quantization function. From the result in the Fig. \ref{fig:exp}, it can be found that the highest accuracy is achieved when $l$ is set to 6. This indicates that historical gradient memory requires an appropriate length, and shorter memory lengths cannot accumulate enough sequence information. A longer memory length introduces more historical gradient noise, which is not conducive to the generation of gradient momentum.

\subsection{Comparison with SOTA Methods}
In order to verify the performance of FSG, we conducted performance studies on it compared to other BNN optimization methods. We follow the experimental setup of \cite{chen2019metaquant,qin2020forward}.  A series of SOTA methods are compared on the CIFAR-10 and CIFAR-100 datasets to validate performance, including DoReFa \cite{zhou2016dorefa}, LSTMFC \cite{chen2019metaquant}, ReSTE \cite{wu2023estimator}, RBNN \cite{lin2020rotated}, IR-Net \cite{qin2020forward}. The experimental results are shown in Tab. \ref{sota}. From the table, it can be seen that our method performs well, surpassing all other methods in terms of final accuracy. On CIFAR-100, it exceeds the second-best baseline, IR-Net, by 0.54\%. 

\begin{table}[t]
    \centering
    \scriptsize
    \renewcommand{\arraystretch}{0.64}
    \caption{Performance comparison with SOTA methods. FP is the full-precision version of the backbone. The bold font is utilized to denote the optimal results, the underline is applied to indicate the suboptimal outcomes.}
    \begin{tabular}{lccc}
         \toprule
         \textbf{Dataset} & \textbf{CIFAR-10} & \textbf{CIFAR-10} & \textbf{CIFAR-100} \\
         \cmidrule{1-4}
         \textbf{Backbone} & \textbf{ResNet-20} & \textbf{ResNet-44} & \textbf{ResNet-56} \\
         \midrule
         FP & 91.70 & 93.55 & 71.22 \\
         DoReFa & 87.96 $\pm$ 0.173 & 90.21 $\pm$ 0.075 & 64.78 $\pm$ 0.533 \\
         ReSTE & 89.26 $\pm$ 0.884 & 90.25 $\pm$ 0.783 & 65.59 $\pm$ 0.733 \\
         LSTMFC & 90.03 $\pm$ 0.299 & 91.91 $\pm$ 0.538 & 66.48 $\pm$ 0.808 \\
         IR-Net & 90.80 $\pm$ 0.971 & 92.37 $\pm$ 0.532 & \underline{68.94 $\pm$ 0.967} \\
         RBNN & 90.89 $\pm$ 0.921 & 91.59 $\pm$ 0.874 & 67.10 $\pm$ 0.933 \\
         \rowcolor{gray!20} \textbf{FSG(ours)} & \textbf{91.00} $\pm$ \textbf{0.804} & \textbf{92.78} $\pm$ \textbf{0.053} & \textbf{69.48} $\pm$ \textbf{0.979} \\
         \bottomrule
    \end{tabular}
    \label{sota}
\end{table}

\section{Conclusion}
This work investigated the generation of non-differentiable function gradients based on hypernetwork methods in BNN optimization. We re-examined the process of gradient generation from the inspiration of momentum, modeled the historical gradient sequence to generate the first-order momentum required for optimization, and introduced the FSG method, using a combination of gradient and momentum to generate the final gradient. In addition, we had added LRE to the slow-net to help generate layer specific fine gradients. The experiments on the CIFAR-10 and CIFAR-100 datasets showed that our method has faster convergence speed, lower loss values, and better performance than competitive baselines.

\bibliography{aaai25}

\begin{thebibliography}{30}
\providecommand{\natexlab}[1]{#1}

\bibitem[{Ajanthan et~al.(2021)Ajanthan, Gupta, Torr, Hartley, and Dokania}]{ajanthan_mirror_2021}
Ajanthan, T.; Gupta, K.; Torr, P.; Hartley, R.; and Dokania, P. 2021.
\newblock Mirror descent view for neural network quantization.
\newblock In \emph{International conference on artificial intelligence and statistics}, 2809--2817. PMLR.

\bibitem[{Andrychowicz et~al.(2016)Andrychowicz, Denil, Gomez, Hoffman, Pfau, Schaul, Shillingford, and De~Freitas}]{andrychowicz2016learning}
Andrychowicz, M.; Denil, M.; Gomez, S.; Hoffman, M.~W.; Pfau, D.; Schaul, T.; Shillingford, B.; and De~Freitas, N. 2016.
\newblock Learning to learn by gradient descent by gradient descent.
\newblock \emph{Advances in neural information processing systems}, 29.

\bibitem[{Banner et~al.(2018)Banner, Hubara, Hoffer, and Soudry}]{banner_scalable_2018}
Banner, R.; Hubara, I.; Hoffer, E.; and Soudry, D. 2018.
\newblock Scalable methods for 8-bit training of neural networks.
\newblock \emph{Advances in neural information processing systems}, 31.

\bibitem[{Chen et~al.(2020)Chen, Wang, Xu, Shi, Xu, Tian, and Xu}]{chen_addernet_2020}
Chen, H.; Wang, Y.; Xu, C.; Shi, B.; Xu, C.; Tian, Q.; and Xu, C. 2020.
\newblock AdderNet: Do we really need multiplications in deep learning?
\newblock In \emph{Proceedings of the IEEE/CVF conference on computer vision and pattern recognition}, 1468--1477.

\bibitem[{Chen, Wang, and Pan(2019)}]{chen2019metaquant}
Chen, S.; Wang, W.; and Pan, S.~J. 2019.
\newblock Metaquant: Learning to quantize by learning to penetrate non-differentiable quantization.
\newblock \emph{Advances in Neural Information Processing Systems}, 32.

\bibitem[{Courbariaux, Bengio, and David(2015)}]{courbariaux2015binaryconnect}
Courbariaux, M.; Bengio, Y.; and David, J.-P. 2015.
\newblock Binaryconnect: Training deep neural networks with binary weights during propagations.
\newblock \emph{Advances in neural information processing systems}, 28.

\bibitem[{Courbariaux et~al.(2016)Courbariaux, Hubara, Soudry, El-Yaniv, and Bengio}]{courbariaux2016binarized}
Courbariaux, M.; Hubara, I.; Soudry, D.; El-Yaniv, R.; and Bengio, Y. 2016.
\newblock Binarized neural networks: Training deep neural networks with weights and activations constrained to+ 1 or-1.
\newblock \emph{arXiv preprint arXiv:1602.02830}.

\bibitem[{Denton et~al.(2014)Denton, Zaremba, Bruna, LeCun, and Fergus}]{denton_exploiting_2014}
Denton, E.~L.; Zaremba, W.; Bruna, J.; LeCun, Y.; and Fergus, R. 2014.
\newblock Exploiting linear structure within convolutional networks for efficient evaluation.
\newblock \emph{Advances in neural information processing systems}, 27.

\bibitem[{Devlin et~al.(2018)Devlin, Chang, Lee, and Toutanova}]{devlin2018bert}
Devlin, J.; Chang, M.-W.; Lee, K.; and Toutanova, K. 2018.
\newblock Bert: Pre-training of deep bidirectional transformers for language understanding.
\newblock \emph{arXiv preprint arXiv:1810.04805}.

\bibitem[{Ding et~al.(2021)Ding, Hao, Tan, Liu, Han, Guo, and Ding}]{ding_resrep_2021}
Ding, X.; Hao, T.; Tan, J.; Liu, J.; Han, J.; Guo, Y.; and Ding, G. 2021.
\newblock Resrep: Lossless cnn pruning via decoupling remembering and forgetting.
\newblock In \emph{Proceedings of the IEEE/CVF international conference on computer vision}, 4510--4520.

\bibitem[{Everingham et~al.(2015)Everingham, Eslami, Van~Gool, Williams, Winn, and Zisserman}]{everingham_pascal_2015}
Everingham, M.; Eslami, S.~A.; Van~Gool, L.; Williams, C.~K.; Winn, J.; and Zisserman, A. 2015.
\newblock The pascal visual object classes challenge: A retrospective.
\newblock \emph{International journal of computer vision}, 111: 98--136.

\bibitem[{Girshick et~al.(2014)Girshick, Donahue, Darrell, and Malik}]{girshick_rich_2014}
Girshick, R.; Donahue, J.; Darrell, T.; and Malik, J. 2014.
\newblock Rich feature hierarchies for accurate object detection and semantic segmentation.
\newblock In \emph{Proceedings of the IEEE conference on computer vision and pattern recognition}, 580--587.

\bibitem[{Gu and Dao(2023)}]{mamba}
Gu, A.; and Dao, T. 2023.
\newblock Mamba: Linear-Time Sequence Modeling with Selective State Spaces.
\newblock \emph{arXiv preprint arXiv:2312.00752}.

\bibitem[{Gu, Goel, and R\'e(2022)}]{gu2022efficiently}
Gu, A.; Goel, K.; and R\'e, C. 2022.
\newblock Efficiently Modeling Long Sequences with Structured State Spaces.
\newblock In \emph{The International Conference on Learning Representations ({ICLR})}.

\bibitem[{Hayashi et~al.(2019)Hayashi, Yamaguchi, Sugawara, and Maeda}]{hayashi_exploring_2019}
Hayashi, K.; Yamaguchi, T.; Sugawara, Y.; and Maeda, S.-i. 2019.
\newblock Exploring unexplored tensor network decompositions for convolutional neural networks.
\newblock \emph{Advances in Neural Information Processing Systems}, 32.

\bibitem[{He et~al.(2016)He, Zhang, Ren, and Sun}]{he_deep_2016}
He, K.; Zhang, X.; Ren, S.; and Sun, J. 2016.
\newblock Deep residual learning for image recognition.
\newblock In \emph{Proceedings of the IEEE conference on computer vision and pattern recognition}, 770--778.

\bibitem[{Hochreiter and Schmidhuber(1997)}]{6795963}
Hochreiter, S.; and Schmidhuber, J. 1997.
\newblock Long Short-Term Memory.
\newblock \emph{Neural Computation}, 9(8): 1735--1780.

\bibitem[{Iandola et~al.(2016)Iandola, Han, Moskewicz, Ashraf, Dally, and Keutzer}]{iandola_squeezenet_2016}
Iandola, F.~N.; Han, S.; Moskewicz, M.~W.; Ashraf, K.; Dally, W.~J.; and Keutzer, K. 2016.
\newblock SqueezeNet: AlexNet-level accuracy with 50x fewer parameters and< 0.5 MB model size.
\newblock \emph{arXiv preprint arXiv:1602.07360}.

\bibitem[{Krizhevsky, Hinton et~al.(2009)}]{krizhevsky2009learning}
Krizhevsky, A.; Hinton, G.; et~al. 2009.
\newblock Learning multiple layers of features from tiny images.

\bibitem[{LeCun, Bengio, and Hinton(2015)}]{lecun_deep_2015}
LeCun, Y.; Bengio, Y.; and Hinton, G. 2015.
\newblock Deep learning.
\newblock \emph{nature}, 521(7553): 436--444.

\bibitem[{Lin et~al.(2020)Lin, Ji, Xu, Zhang, Wang, Wu, Huang, and Lin}]{lin2020rotated}
Lin, M.; Ji, R.; Xu, Z.; Zhang, B.; Wang, Y.; Wu, Y.; Huang, F.; and Lin, C.-W. 2020.
\newblock Rotated binary neural network.
\newblock \emph{Advances in neural information processing systems}, 33: 7474--7485.

\bibitem[{Liu et~al.(2020)Liu, Wen, Wang, Tao, Chen, Osa, and Kato}]{liu2020quantnet}
Liu, J.; Wen, D.; Wang, D.; Tao, W.; Chen, T.-W.; Osa, K.; and Kato, M. 2020.
\newblock QuantNet: Learning to quantize by learning within fully differentiable framework.
\newblock In \emph{Computer Vision--ECCV 2020 Workshops: Glasgow, UK, August 23--28, 2020, Proceedings, Part V 16}, 38--53. Springer.

\bibitem[{Luo, Wu, and Lin(2017)}]{luo_thinet_2017}
Luo, J.-H.; Wu, J.; and Lin, W. 2017.
\newblock Thinet: A filter level pruning method for deep neural network compression.
\newblock In \emph{Proceedings of the IEEE international conference on computer vision}, 5058--5066.

\bibitem[{Qin et~al.(2020)Qin, Gong, Liu, Shen, Wei, Yu, and Song}]{qin2020forward}
Qin, H.; Gong, R.; Liu, X.; Shen, M.; Wei, Z.; Yu, F.; and Song, J. 2020.
\newblock Forward and backward information retention for accurate binary neural networks.
\newblock In \emph{Proceedings of the IEEE/CVF conference on computer vision and pattern recognition}, 2250--2259.

\bibitem[{Rastegari et~al.(2016)Rastegari, Ordonez, Redmon, and Farhadi}]{rastegari2016xnor}
Rastegari, M.; Ordonez, V.; Redmon, J.; and Farhadi, A. 2016.
\newblock Xnor-net: Imagenet classification using binary convolutional neural networks.
\newblock In \emph{European conference on computer vision}, 525--542. Springer.

\bibitem[{Saad(1998)}]{saad1998online}
Saad, D. 1998.
\newblock Online algorithms and stochastic approximations.
\newblock \emph{Online Learning}, 5(3): 6.

\bibitem[{Wu et~al.(2023)Wu, Zheng, Liu, and Zheng}]{wu2023estimator}
Wu, X.-M.; Zheng, D.; Liu, Z.; and Zheng, W.-S. 2023.
\newblock Estimator meets equilibrium perspective: A rectified straight through estimator for binary neural networks training.
\newblock In \emph{Proceedings of the IEEE/CVF International Conference on Computer Vision}, 17055--17064.

\bibitem[{Xu et~al.(2021)Xu, Han, Xu, Tang, Xu, and Wang}]{xu2021learning}
Xu, Y.; Han, K.; Xu, C.; Tang, Y.; Xu, C.; and Wang, Y. 2021.
\newblock Learning frequency domain approximation for binary neural networks.
\newblock \emph{Advances in Neural Information Processing Systems}, 34: 25553--25565.

\bibitem[{Zhou et~al.(2016)Zhou, Wu, Ni, Zhou, Wen, and Zou}]{zhou2016dorefa}
Zhou, S.; Wu, Y.; Ni, Z.; Zhou, X.; Wen, H.; and Zou, Y. 2016.
\newblock Dorefa-net: Training low bitwidth convolutional neural networks with low bitwidth gradients.
\newblock \emph{arXiv preprint arXiv:1606.06160}.

\bibitem[{Zhu et~al.(2024)Zhu, Liao, Zhang, Wang, Liu, and Wang}]{vim}
Zhu, L.; Liao, B.; Zhang, Q.; Wang, X.; Liu, W.; and Wang, X. 2024.
\newblock Vision Mamba: Efficient Visual Representation Learning with Bidirectional State Space Model.
\newblock \emph{arXiv preprint arXiv:2401.09417}.

\end{thebibliography}

\section{Reproducibility Checklist}

This paper
\begin{itemize}
\item Includes a conceptual outline and/or pseudocode description of AI methods introduced (\textbf{yes})
\item Clearly delineates statements that are opinions, hypothesis, and speculation from objective facts and results (\textbf{yes})
\item Provides well marked pedagogical references for less-familiare readers to gain background necessary to replicate the paper (\textbf{yes})
\end{itemize}
\textbf{Does this paper make theoretical contributions?} (\textbf{yes})

If yes, please complete the list below.
\begin{itemize}
\item All assumptions and restrictions are stated clearly and formally. (\textbf{yes})
\item All novel claims are stated formally (e.g., in theorem statements). (\textbf{yes})
\item Proofs of all novel claims are included. (\textbf{yes})
\item Proof sketches or intuitions are given for complex and/or novel results. (\textbf{yes})
\item Appropriate citations to theoretical tools used are given. (\textbf{yes})
\item All theoretical claims are demonstrated empirically to hold. (\textbf{yes})
\item All experimental code used to eliminate or disprove claims is included. (\textbf{NA})
\end{itemize}
\textbf{Does this paper rely on one or more datasets? (yes)}

If yes, please complete the list below.
\begin{itemize}
\item A motivation is given for why the experiments are conducted on the selected datasets (\textbf{yes})
\item All novel datasets introduced in this paper are included in a data appendix. (\textbf{NA})
\item All novel datasets introduced in this paper will be made publicly available upon publication of the paper with a license that allows free usage for research purposes. (\textbf{NA})
\item All datasets drawn from the existing literature (potentially including authors’ own previously published work) are accompanied by appropriate citations. (\textbf{yes})
\item All datasets drawn from the existing literature (potentially including authors’ own previously published work) are publicly available. (\textbf{yes})
\item All datasets that are not publicly available are described in detail, with explanation why publicly available alternatives are not scientifically satisficing. (\textbf{NA})
\end{itemize}
\textbf{Does this paper include computational experiments?} (\textbf{yes})

If yes, please complete the list below.
\begin{itemize}
\item Any code required for pre-processing data is included in the appendix. (\textbf{no}).
\item All source code required for conducting and analyzing the experiments is included in a code appendix. (\textbf{no})
\item All source code required for conducting and analyzing the experiments will be made publicly available upon publication of the paper with a license that allows free usage for research purposes. (\textbf{yes})
\item All source code implementing new methods have comments detailing the implementation, with references to the paper where each step comes from (\textbf{partial})
\item If an algorithm depends on randomness, then the method used for setting seeds is described in a way sufficient to allow replication of results. (\textbf{yes})
\item This paper specifies the computing infrastructure used for running experiments (hardware and software), including GPU/CPU models; amount of memory; operating system; names and versions of relevant software libraries and frameworks. (\textbf{partial})
\item This paper formally describes evaluation metrics used and explains the motivation for choosing these metrics. (\textbf{yes})
\item This paper states the number of algorithm runs used to compute each reported result. (\textbf{yes})
\item Analysis of experiments goes beyond single-dimensional summaries of performance (e.g., average; median) to include measures of variation, confidence, or other distributional information. (\textbf{yes})
\item The significance of any improvement or decrease in performance is judged using appropriate statistical tests (e.g., Wilcoxon signed-rank). (\textbf{no})
\item This paper lists all final (hyper-)parameters used for each model/algorithm in the paper’s experiments. (\textbf{yes})
\item This paper states the number and range of values tried per (hyper-) parameter during development of the paper, along with the criterion used for selecting the final parameter setting. (\textbf{yes})
\end{itemize}

\newpage

\appendix

\section{More Experiment Setup}
For fair comparison, we set the initial learning rate to $1e^{-3}$ for CIFAR-10 dataset and $1e^{-1}$ for CIFAR-100 dataset, and set the epoch of all experiments to 100. After every 30 epochs, the learning rate will decay to the $1/10$ of its original value. All experiments are conducted 5 times and the mean and variance of the results are reported. All models are implemented using PyTorch on NVIDIA RTX3090 or NVIDIA RTX A6000 GPU.

For FSG, slow-net uses Mamba block \cite{mamba} with block expansion factor set to 100 and incorporates learnable embedding to recognize layer information. Fast-net uses 3 fully connected layers with hidden dimension set to 100 and no non-linear activation. For both fast-net and slow-net, the Adam optimizer is used for optimization, with an initial learning rate set to $lr=1e^{-3}$. When accumulating gradients, the weight is set to $\alpha=0.3$. For the slow-net, we set the historical gradient memory length to $l=6$.

\section{Computational Overhead Analysis}
In this experiment, we quantify the additional computational overhead caused by introducing fast and slow networks over time. It should be noted that removing the fast network degrades it to the baseline method for comparison. We used ResNet-56 as the backbone on the CIFAR-100 data set to compare the computational overhead increased by introducing fast networks and slow networks on the STE method. From Table \ref{computational} that the introduction of slow-net increases computation time by 3.2 times, however, In real deployment fast-net and slow-net are removed, FSG is able to provide better test performance without any extra inference time.

\begin{table}[h]
    \centering
    \scriptsize
    \caption{Computational Overhead Analysis}
    \begin{tabular}{lllcc}
         \toprule
         \textbf{Method} & \textbf{Fast-net} & \textbf{Slow-net} & \textbf{Seconds/Epoch} & \textbf{Test Acc (\%)} \\
         \midrule
         STE & - & - & \textbf{42.28} & 64.78 \\
         FCGrad & Multi MLP & - & 61.26 & 67.86 \\
         ReSTE & Multi MLP & Mamba & 196.47 & \textbf{69.48} \\
         \bottomrule
    \end{tabular}
    \label{computational}
\end{table}

\section{SSM Based Model and Mamba}

State space models (SSMs) \cite{gu2022efficiently}, originates from modern control system theory and can be transformed into the following equation,
\begin{equation}
    \begin{aligned}
        h'(t)&=\mathbf{A}h(t)+\mathbf{B}x(t),\\
        y(t)&=\mathbf{C}h(t),
    \end{aligned}
    \label{ssm-model}
\end{equation}
where $x(t) \in \mathbb{R}$ represents the input sequence, $h(t) \in \mathbb{R}^N$ represents the hidden state and $y(t)\in \mathbb{R}$ represents the predicted output sequence. It is a model used to describe the state representation of a sequence at each time step and predict its next state based on input.

The original theory is used to describe continuous functions, and the SSM based models are its discrete version. The continuous parameters $(\mathbf{\Delta}, \mathbf{A}, \mathbf{B})$ are transformed into discrete parameters $(\overline{\mathbf{A}}, \overline{\mathbf{B}})$, and the commonly used discretization rule is zero-order hold (ZOH),
\begin{equation}
    \overline{\mathbf{A}}=\exp(\mathbf{\Delta} \mathbf{A}),\quad\overline{\mathbf{B}}=(\mathbf{\Delta} \mathbf{A})^{-1}(\exp(\mathbf{\Delta} \mathbf{A})-I)\cdot\mathbf{\Delta} \mathbf{B},
    \label{dis-rule}
\end{equation}
After the discretization of $\mathbf{A}$, $\mathbf{B}$, the discretized version of Eq. \ref{ssm-model} using a step size $\mathbf{\Delta}$ can be rewritten as,
\begin{equation}
    \begin{aligned}
        &h_t=\overline{\mathbf{A}}h_{t-1}+\overline{\mathbf{B}}x_t,\\
        &y_t=\mathbf{C}h_t,
    \end{aligned}
    \label{discretized}
\end{equation}
For discrete versions of the model, Eq. \ref{discretized} can be used for linear recursive calculation. And global convolution Eq. \ref{global-conv} can be applied for convolutional computation, 
\begin{equation}
    \begin{aligned}
        &\overline{\mathbf{K}}=(\mathbf{C}\overline{\mathbf{B}},\mathbf{C}\overline{\mathbf{A}}\overline{\mathbf{B}},...,\mathbf{C}\overline{\mathbf{A}}^{L-1}\overline{\mathbf{B}}),\\
        &y=x*\overline{\mathbf{K}},
    \end{aligned}
    \label{global-conv}
\end{equation}
where $\overline{\mathbf{K}}$ is a definite structured convolutional kernel. 

Mamba \cite{mamba} incorporates a selection mechanism into the model by incorporating input dependent parameters that affect sequence interaction, which turns some parameters $(\mathbf{\Delta}, \mathbf{B}, \mathbf{C})$ with functions of input sequence $x$ of batch size $B$ and length $L$ with $D$ channels, $x \in \mathbb{R}^{B\times L\times D}$,
\begin{equation}
    \begin{aligned}
        &\mathbf{B} \in \mathbb{R}^{B \times L \times N}\leftarrow s_{\mathbf{B}}(x)=\mathrm{Linear}_N(x),\\
        &\mathbf{C} \in \mathbb{R}^{B \times L \times N}\leftarrow s_{\mathbf{C}}(x)=\mathrm{Linear}_N(x),\\
        &\mathbf{\Delta} \in \mathbb{R}^{B \times L \times D}\leftarrow s_{\mathbf{\Delta}}(x)=\mathrm{Broadcast}_D(\mathrm{Linear}_1(x)),
    \end{aligned}
\end{equation}

where $\mathrm{Linear}_d$ is a parameterized projection to dimension $d$. $\overline{\mathbf{A}}$ and $\overline{\mathbf{B}}$ are generated by input-driven generation through Eq. \ref{dis-rule}.

The SSM model is characterized by a state equation that captures the temporal evolution of the system's state, and an observation equation that models the output based on the current state and input. Therefore, SSM based models are suitable for processing sequence data. By mapping time-series data to state space, SSM based models can more effectively capture long-term dependencies, which is crucial for understanding and predicting gradient changes over time. 
Combining the advantages of transformer parallel training and RNN linear inference ability, SSM based models are suitable for processing massive gradient information. 
Due to the gradients in the early stages of optimization often containing a significant amount of noise, and SSM-based models being more prone to noise accumulation due to their fixed propagation patterns, further selection of historical gradients is necessary.

\section{Algorithm for FSG}

\begin{algorithm}[h]
    \caption{Training Stage of FSG}          
    \begin{algorithmic}[1]
    \REQUIRE Training dataset $\{\mathbf{x}, \mathbf{y}\}^N$, well-trained full precision base model $\mathbf{W}$.
    \ENSURE Quantized base model $\mathbf{W}_b$
    \STATE{Construct shared hypernetwork $\mathcal{M}_f, \mathcal{M}_s$, training iteration $t=1$.}
    \WHILE{not optimal}
    \FOR{Layer $i$ from $1$ to $n$}
    \STATE{$(W_i^{t+1})_b=Q\{\mathcal{A}[W_i^{t}-\alpha\mathcal{M}_f(\mathcal{G}_{W_i}^{t}, \hat{W}_i^{t})\frac{\partial \mathcal{A}(W_i^t)}{\partial W_i^t}+\beta\mathcal{M}_s(\hbar_{t})] \}.$}
    \ENDFOR
    \STATE{Calculate loss: $\ell=\text{Loss}\left\{(f\left[Q(\mathcal{A}(\mathcal{W})); \mathbf{x}\right], \mathbf{y}\right\}$}.
    \STATE{Generate $g_{W^t}$ using chain rules.}
    \STATE{Calculate fast gradient $g_f=\mathcal{M}_f(\mathcal{G}_{W_i}^{t}, \hat{W}_i^{t})$.}
    \STATE{Calculate slow gradient $g_s=\mathcal{M}_s(\hbar_{t})$.}
    \STATE{Calculate the final gradient: $\mathcal{G}_{FSG}=\beta g_s -\alpha g_f\frac{\partial \mathcal{A}(W_i^t)}{\partial W_i^t}$.}
    \STATE{Calculate $\frac{\partial\ell}{\partial\phi_{f}^{t}}$ by \ref{gradient_f}.}
    \STATE{Calculate $\frac{\partial\ell}{\partial\phi_{s}^{t}}$ by \ref{gradient_s}.}
    \FOR{Layer $i$ from $1$ to $n$}
    \STATE{$\mathbf{W}_i^{t+1}=\mathbf{W}_i^t-\alpha\mathcal{M}_f(\mathcal{G}_{W_i}^{t}, \hat{W}_i^{t})+\beta\mathcal{M}_s(\hbar_{t}).$}
    \ENDFOR
    \STATE{$\phi_f^{t+1}=\phi_f^t-\gamma_f*\frac{\partial\ell}{\partial\phi_{f}^{t}}$($\gamma_f$ is the learning rate of the fast-net).}
    \STATE{$\phi_s^{t+1}=\phi_s^t-\gamma_s*\frac{\partial\ell}{\partial\phi_{s}^{t}}$($\gamma_s$ is the learning rate of the slow-net).}
    \STATE{$t=t+1$.}
    \ENDWHILE
    \end{algorithmic}
    \label{Algo1} 
\end{algorithm}

The detailed process of the fast and slow gradient generation mechanism is described in the algorithm \ref{Algo1}. First, two shared hypernetwork $\mathcal{M}_f, \mathcal{M}_s$ are constructed, where the slow-net is randomly initialized and the fast-net is initialized with a random orthogonal matrix. 

In each training iteration, lines 2-6 describe the forward process: for each layer of slow-net, $\hbar_{t}$ from the previous iteration is received to generate the gradient momentum, and $g_{W_i}^{t}$ and $\hat{W}_i^{t}$ from the previous iteration are input into $\mathcal{M}_f$ to generate the current gradient, finally the final gradient \(\mathcal{G}_{FSG}\) is generated by combining the momentum for inference (as shown in lines 3-5). Since $g_{W}$ is not calculated in the first iteration, normal quantization training is performed in the first iteration, i.e. the fourth line is replaced by $(\mathbf{W}_i^1)_b=Q(\hat{\mathbf{W}}_i^0)=Q\left[\mathcal{A}(\mathbf{W}_i^0)\right]$. 

Lines 7-12 show the back propagation process: the gradient of $\mathbf{W}^t$ can be obtained by back propagation (as shown in the seventh line). In the back propagation process, the $g_{W_i}^{t}$ and $\hat{W}_i^{t}$ of the current iteration are obtained, and the $\hbar_{i}^{t}$ of the current iteration is obtained from HGS, which are input into $\mathcal{M}_f$ and $\mathcal{M}_s$ respectively. The final gradient \(\mathcal{G}_{FSG}\) is saved to the next iteration for calculation, that is, $g_{W_i}^{t+1}$ (as shown in lines 8-10). 

$\mathcal{M}_f$ and $\mathcal{M}_s$ are included in the inference graph to obtain their gradients (as shown in lines 11-12). Finally, the registered gradient \(\mathcal{G}_{FSG}\) is used to update each layer of the network (as shown in lines 13-15). 
Note that in the first iteration, no gradient is generated, so the weights are not updated. 
Finally, the slow-net and fast-net are updated (as shown in lines 16-17).

\section{Proof of Theorem 1}

Before proving this, we first make the following assumptions,
\begin{itemize}
    \item Assume that each \(f_i(x)\) is a closed convex function, and there is a subgradient, that is, \(g\) is the subgradient of \(f\) at \(x\). If for \(\forall y, f(y)\geq f(x)+g^\top(y-x)\), define \(\partial f(x)\) as a set of subgradients;
    \item Define \(\mathcal{G}_k=\mathcal{G}(x_k;\xi_k)\) as the stochastic gradient of \(f(x)\) at \(x_k\) depending on a random variable \(\xi_k\), and the expectation of the stochastic subgradient at \(x_k\) is the gradient at \(x_k\), that is, \(\mathrm{E}[\mathcal{G}(\mathbf{x}_k;\xi_k)]=\nabla f(\mathbf{x}_k)\), assuming that the variance of the stochastic subgradient is bounded, that is, \(\mathbb{E}[\|\mathcal{G}(\mathbf{x};\xi)-\mathbb{E}[\mathcal{G}(\mathbf{x};\xi)]\|^2]\leq\delta^2\);
    \item Assume that the second-order moment of the random subgradient is uniformly bounded, that is, \(\exists G\), for \(\forall x \in \mathbb{R}^n\), \(\|\partial f(x)\|\leq G\);
    \item Assume that the iterated random point sequence \(\{\mathbf{x}_k\}\) is bounded everywhere, that is, \(\exists \kappa, \exists \rho\) has \(\kappa\leq \|\mathbf{x}_k-\mathbf{x}^*\|\leq\rho\) for \(\forall k \), where \(\mathbf{x}^*\) is the optimal solution to the problem (\ref{formual:question});
    \item Assuming that the momentum generated by the slow network is random and the expected error between it and the actual gradient momentum is zero, that is, \(\mathbb{E}_k(\Delta_s^k)=\mathbb{E}_k[(\mathcal{M}_s(\hbar_k)-(x_k-x_{k-1}))]=0\);
    \item Assuming that the multi-layer fully connected layer \(\phi_f\) of fast-net is initialized as a positive square matrix, it is still bounded after mapping the bounded point column, that is, \(\exists \Omega, \exists \Theta \), and for \(\forall x_k\), \(\Omega\|x_k\| \leq \|\phi_f(x_k)\|\leq \Theta\|x_k\|\).
\end{itemize}
Then, we give a lemma,
\begin{lemma}
    formula \ref{corporate} contains the following recursion,
    \begin{equation}
        \begin{aligned}
            \mathbf{x}_{k+1}+\mathbf{p}_{k+1}= &\mathbf{x}_k+\mathbf{p}_k-\frac\alpha{1-\beta}\mathcal{M}_f(\mathcal{G}(\mathbf{x}_k), x_k)\\
            &+\frac{\beta}{1-\beta}\Delta_s^k,\quad k\geq0
        \end{aligned}
        \label{formual:gengxing}
    \end{equation}
    \begin{equation}
        \mathbf{v}_{k+1}=\beta\mathbf{v}_k-\alpha\mathcal{M}_f(\mathcal{G}(\mathbf{x}_k), x_k),\quad k\geq0
    \end{equation}
    where \(\mathbf{v}_{k}=\frac{1-\beta}{\beta}\mathbf{p}_{k}\), \(\mathbf{p}_{k}\) is:
    \begin{equation}
        \mathbf{p}_k=\begin{cases}\frac\beta{1-\beta}(x_k-x_{k-1}),\quad k\geq1,\\0,\quad k=0.&\end{cases}
        \label{formual:pk}
    \end{equation}
    Let \(\mathbf{x}_{-1}=\mathbf{x}_0\), For \(\forall k \geq 0\), we have,
    \begin{equation}
        \begin{aligned}
            &\mathbb{E}[\|\mathbf{x}_{k+1}+\mathbf{p}_{k+1}-\mathbf{x}\|^{2}] \leq\mathbb{E}[\|\mathbf{x}_{k}+\mathbf{p}_{k}-\mathbf{x}\|^{2}] \\
            &-\frac{2\alpha\Omega\kappa}{1-\beta}\mathbb{E}[(f(\mathbf{x}_{k})-f(\mathbf{x}))]  \\
            &-\frac{2\alpha\beta\Omega\kappa}{(1-\beta)^{2}}\mathbb{E}[(f(\mathbf{x}_{k})-f(\mathbf{x}_{k-1}))] \\
            &+\left(\frac{\alpha\Theta\rho}{1-\beta}\right)^{2}(G^{2}+\delta^{2})\quad k\ge1.
        \end{aligned}
        \label{formual:lemma1}
    \end{equation}
\end{lemma}
The lemma states that the expected error of the recursive function decreases during the iterations. Then we give the following theorem based on the lemma.
\begin{proof}
    From the recursive relation in formula (\ref{formual:gengxing}) and the definition of \(\mathbf{p}_{k}\) in formula (\ref{formual:pk}), we know that for \(\forall x \in \mathbb{R}^{d}\) and \( k\geq1\),
    \begin{equation*}
        \begin{aligned}
            &\|\mathbf{x}_{k+1}+\mathbf{p}_{k+1}-\mathbf{x}\|^2 \\
            =&\|\mathbf{x}_k+\mathbf{p}_k-\frac{\alpha}{1-\beta}\mathcal{M}_f(\mathcal{G}_k, x_k)+\frac{\beta}{1-\beta}\Delta_s^k-\mathbf{x}\|^2 \\
            =&\|\mathbf{x}_k+\mathbf{p}_k-\mathbf{x}\|^2-\frac{2\alpha}{1-\beta}(\mathbf{x}_k+\mathbf{p}_k-\mathbf{x})^\top\mathcal{M}_f(\mathcal{G}_k, x_k) \\
            &+\left(\frac{\alpha}{1-\beta}\right)^2\|\mathcal{M}_f(\mathcal{G}_k, x_k)\|^2 + \left(\frac{\beta}{1-\beta}\right)^2{\Delta_s^k}^2\\
            &+\frac{2\beta}{1-\beta}\Delta_s^k(\mathbf{x}_k+\mathbf{p}_k-\mathbf{x})^\top-\frac{2\alpha\beta}{(1-\beta)^2}\mathcal{M}_f(\mathcal{G}_k, x_k)\Delta_s^k\\
            =&\|\mathbf{x}_k+\mathbf{p}_k-\mathbf{x}\|^2-\frac{2\alpha}{1-\beta}(\mathbf{x}_k+\mathbf{p}_k-\mathbf{x})^\top\phi_f(x_k)\mathcal{G}_k \\
            &+\left(\frac{\alpha}{1-\beta}\right)^2\|\phi_f(x_k)\mathcal{G}_k\|^2 + \Delta_s^k \cdot \ast\\
            =&\|\mathbf{x}_{k}+\mathbf{p}_{k}-\mathbf{x}\|^{2}-\frac{2\alpha}{1-\beta}(\mathbf{x}_{k}-\mathbf{x})^{\top}\phi_f(x_k)\mathcal{G}_k \\
            &-\frac{2\alpha\beta}{(1-\beta)^{2}}\mathcal{M}_s(\hbar_k)\phi_f(x_k)\mathcal{G}_k \\
            &+\left(\frac{\alpha}{1-\beta}\right)^{2}\|\phi_f(x_k)\mathcal{G}_k\|^{2} + \Delta_s^k \cdot \ast\\
            =&\|\mathbf{x}_{k}+\mathbf{p}_{k}-\mathbf{x}\|^{2}-\frac{2\alpha}{1-\beta}(\mathbf{x}_{k}-\mathbf{x})^{\top}\phi_f(x_k)(\delta_k+\partial f(\mathbf{x}_k)) \\
            &-\frac{2\alpha\beta}{(1-\beta)^{2}}\phi_s(\mathcal{G}_k)\phi_f(x_k)(\delta_k+\partial f(\mathbf{x}_k))\\
            &+\left(\frac{\alpha}{1-\beta}\right)^{2}\|\phi_f(x_k)(\delta_k+\partial f(\mathbf{x}_k))\|^{2}+ \Delta_s^k \cdot \ast \\
        \end{aligned}
    \end{equation*}
    Where \(\ast=\left(\frac{\beta}{1-\beta}\right)^2{\Delta_s^k} +\frac{2\beta}{1-\beta}(\mathbf{x}_k+\mathbf{p}_k-\mathbf{x})^\top-\frac{2\alpha\beta}{(1-\beta)^2}\mathcal{M}_f(\mathcal{G}_k, x_k)\).Both sides of the equation take the expectation of randomness in \(\xi_1, \xi_2, \dots, \xi_k\),
    \begin{equation}
        \begin{aligned}
            &\mathbb{E}_k[\|\mathbf{x}_{k+1}+\mathbf{p}_{k+1}-\mathbf{x}\|^2]=\mathbb{E}_k[\|\mathbf{x}_{k}+\mathbf{p}_{k}-\mathbf{x}\|^{2}] \\
            &-\frac{2\alpha}{1-\beta}\mathbb{E}_k[(\mathbf{x}_{k}-\mathbf{x})^{\top}\phi_f(x_k)(\delta_k+\partial f(\mathbf{x}_k))] \\
            &-\frac{2\alpha\beta}{(1-\beta)^{2}}\mathbb{E}_k[\phi_s(\mathcal{G}_k)\phi_f(x_k)(\delta_k+\partial f(\mathbf{x}_k))] \\
            &+\left(\frac{\alpha}{1-\beta}\right)^{2}\mathbb{E}_k[\|\phi_f(x_k)(\delta_k+\partial f(\mathbf{x}_k))\|^{2}] \\
        \end{aligned}
        \label{formual:1}
    \end{equation}
    Also note that,
    \begin{equation*}
        \begin{aligned}
            &\mathbb{E}_k[(\mathbf{x}_k-\mathbf{x})^\top\phi_f(x_k)(\delta_k+\partial f(\mathbf{x}_k))] \\
            =&\mathbb{E}_k[\phi_f(x_k)]\mathbb{E}_{k-1}[(\mathbf{x}_k-\mathbf{x})^\top\partial f(\mathbf{x}_k)]
        \end{aligned}
    \end{equation*}
    \begin{equation*}
        \begin{aligned}
            &\mathbb{E}_k[\phi_s(\mathcal{G}_k)\phi_f(x_k)(\delta_k+\partial f(\mathbf{x}_k))] \\
            =&\mathbb{E}_k[\phi_s(\mathcal{G}_k)]\mathbb{E}_[\phi_f(x_k)]\mathbb{E}_k[(\delta_k+\partial f(\mathbf{x}_k))] \\
            =&\mathbb{E}_k[\phi_f(x_k)]\mathbb{E}_k[(\mathbf{x}_k-\mathbf{x}_{k-1})^\top(\delta_k+\partial f(\mathbf{x}_k))] \\
            =&\mathbb{E}_k[\phi_f(x_k)]\mathbb{E}_{k-1}[(\mathbf{x}_k-\mathbf{x}_{k-1})^\top\partial f(\mathbf{x}_k)]\\
        \end{aligned}
    \end{equation*}
    \begin{equation*}
        \begin{aligned}
            &\mathbb{E}_k[\|\phi_f(x_k)(\delta_k+\partial f(\mathbf{x}_k))\|^{2}] \\
            =&\mathbb{E}_k[\|\phi_f(x_k)\|^2]\mathbb{E}_k[\|\delta_k+\partial f(\mathbf{x}_k)\|^2] \\
            =&\mathbb{E}_k[\|\phi_f(x_k)\|^2](\mathbb{E}_k[\|\delta_k\|^2]+\mathbb{E}_{k-1}[\|\partial f(\mathbf{x}_k)\|^2]) \\
        \end{aligned}
    \end{equation*}
    Then the formula (\ref{formual:1}) is,
    \begin{equation*}
        \begin{aligned}
            &\mathbb{E}_k[\|\mathbf{x}_{k+1}+\mathbf{p}_{k+1}-\mathbf{x}\|^2] =\mathbb{E}_{k-1}[\|\mathbf{x}_k+\mathbf{p}_k-\mathbf{x}\|^2] \\
            &-\frac{2\alpha}{1-\beta}\mathbb{E}_k[\phi_f(x_k)]\mathbb{E}_{k-1}[(\mathbf{x}_k-\mathbf{x})^\top\partial f(\mathbf{x}_k)] \\
            &-\frac{2\alpha\beta}{(1-\beta)^2}\mathbb{E}_k[\phi_f(x_k)]\mathbb{E}_{k-1}[(\mathbf{x}_k-\mathbf{x}_{k-1})^\top\partial f(\mathbf{x}_k)] \\
            &+\left(\frac{\alpha}{1-\beta}\right)^2\mathbb{E}_k[\|\phi_f(x_k)\|^2](\mathbb{E}_k[\|\delta_k\|^2] \\
            &+\mathbb{E}_{k-1}[\|\partial f(\mathbf{x}_k)\|^2])
            \end{aligned}
    \end{equation*}
    Note that,
    \begin{equation*}
        \begin{aligned}
            &\begin{aligned}f(\mathbf{x}_k)-f(\mathbf{x})\leq(\mathbf{x}_k-\mathbf{x})^\top\partial f(\mathbf{x}_k),\end{aligned} \\
            &\begin{aligned}f(\mathbf{x}_k)-f(\mathbf{x}_{k-1})\leq(\mathbf{x}_k-\mathbf{x}_{k-1})^\top\partial f(\mathbf{x}_k),\end{aligned} \\
            &\mathbb{E}_{k}[\|\delta_{k}\|^{2}]\leq\delta^{2},\quad\mathbb{E}_{k-1}[\|\partial f(\mathbf{x}_{k})\|^{2}]\leq G^{2}, \\
            &\Omega\kappa \leq \mathbb{E}_k[\phi_f(x_k)]\leq \Theta\rho, \\
            &\Omega^2\kappa^2 \leq \mathbb{E}_k[\|\phi_f(x_k)\|^2]\leq \Theta^2\rho^2, \\
        \end{aligned}
    \end{equation*}
    The first two inequalities are properties of subgradients, and the last two inequalities are caused by the boundedness assumption. Therefore,
    \begin{equation*}
        \begin{aligned}
            &\mathbb{E}[\|\mathbf{x}_{k+1}+\mathbf{p}_{k+1}-\mathbf{x}\|^{2}] \\
            \leq&\mathbb{E}[\|\mathbf{x}_{k}+\mathbf{p}_{k}-\mathbf{x}\|^{2}]-\frac{2\alpha\Omega\kappa}{1-\beta}\mathbb{E}[(f(\mathbf{x}_{k})-f(\mathbf{x}))]  \\
            &-\frac{2\alpha\beta\Omega\kappa}{(1-\beta)^{2}}\mathbb{E}[(f(\mathbf{x}_{k})-f(\mathbf{x}_{k-1}))] \\
            &+\left(\frac{\alpha\Theta\rho}{1-\beta}\right)^{2}(G^{2}+\delta^{2})\quad k\ge1
        \end{aligned}
    \end{equation*}
    By setting \(x_{-1}=x_0\) and similar analysis as above, the above inequality also holds for \(k=0\).
\end{proof}

Next, we use Lemma 2 to prove Theorem 1.
\begin{proof}
    Summing the inequality in Lemma 1 (\ref{formual:lemma1}) from \(k=0,\dots,t\),
    \begin{equation*}
        \begin{aligned}
            &\frac{2\alpha\Omega\kappa}{1-\beta}\sum_{k=0}^t\mathbb{E}[f(\mathbf{x}_k)-f(\mathbf{x})] \\
            \leq&\frac{2\alpha\beta\Omega\kappa}{(1-\beta)^2}(f(\mathbf{x}_0)-f(\mathbf{x}_t))+\|\mathbf{x}_0-\mathbf{x}\|^2 \\
            &+\left(\frac{\alpha\Theta\rho}{1-\beta}\right)^{2}(G^2+\delta^2)(t+1)
        \end{aligned}
    \end{equation*}
    Let \(x=x^*\), and note that \(f(x_t)\geq f(x^*)\), then we have,
    \begin{equation*}
        \begin{aligned}
            &\sum_{k=0}^t\mathbb{E}[f(\mathbf{x}_k)-f(\mathbf{x}^*)] \\
            \leq& \frac{\beta}{1-\beta}(f(\mathbf{x}_0)-f(\mathbf{x}^*)) +\frac{1-\beta}{2\alpha\Omega\kappa}\|\mathbf{x}_0-\mathbf{x}^*\|^2 \\
            &+\frac{\alpha\Theta\rho}{2\Omega\kappa(1-\beta)}(G^2+\delta^2)(t+1)
        \end{aligned}
    \end{equation*}
    Define \(\widehat{\mathbf{x}}_t=\sum_{k=0}^t\mathbf{x}_k/(t+1)\). By the convexity of \(f(x)\),
    \begin{equation*}
        \begin{aligned}
            \mathbb{E}[f(\widehat{\mathbf{x}}_t)-f(\mathbf{x}^*)]\leq&\frac\beta{(1-\beta)(t+1)}(f(\mathbf{x}_0)-f(\mathbf{x}^*)) \\
        &+\frac{(1-\beta)\|\mathbf{x}_0-\mathbf{x}_*\|^2}{2\alpha\Omega\kappa(t+1)} \\
        &+\frac{\alpha\Theta\rho(G^2+\delta^2)}{2\Omega\kappa(1-\beta)}
        \end{aligned}
    \end{equation*}
    Substituting the value of \(\alpha\),
    \begin{equation*}
        \begin{aligned}
            &\mathbb{E}[f(\widehat{\mathbf{x}}_t)-f(\mathbf{x}^*)] \\
            \leq &\frac\beta{(1-\beta)(t+1)}(f(\mathbf{x}_0)-f(\mathbf{x}^*)) +\frac{(1-\beta)\|\mathbf{x}_0-\mathbf{x}_*\|^2}{2C\Omega\kappa\sqrt{t+1}} \\
            &+\frac{C\Theta\rho(G^2+\delta^2)}{2\Omega\kappa(1-\beta)\sqrt{t+1}}\\
        \end{aligned}
    \end{equation*}
\end{proof}

\section{Limitation}
This research focuses on an in-depth discussion of the quantization issues of the ResNet network architecture, especially the quantization strategies of the convolutional layer and linear layer. However, current research does not extend to more complex models such as Transformers and large language models (LLM). In future research, we plan to explore the role of FSG in training these large and complex models, with a view to revealing its potential in the field of network quantification.

\end{document}